\documentclass{article}
\usepackage{amsmath}
\usepackage{graphicx}
\usepackage{caption}
\usepackage{subcaption}
\PassOptionsToPackage{numbers, compress}{natbib}
\usepackage[preprint]{neurips_2026}

\usepackage[utf8]{inputenc} % allow utf-8 input
\usepackage[T1]{fontenc}    % use 8-bit T1 fonts
\usepackage{hyperref}       % hyperlinks
\usepackage{url}            % simple URL typesetting
\usepackage{booktabs}       % professional-quality tables
\usepackage{amsfonts}       % blackboard math symbols
\usepackage{nicefrac}       % compact symbols for 1/2, etc.
\usepackage{microtype}      % microtypography
\usepackage{xcolor}         % colors

\definecolor{ScoreYes}{RGB}{46,125,50}      % green
\definecolor{ScorePart}{RGB}{239,108,0}     % orange
\definecolor{ScoreNo}{RGB}{189,189,189}     % light gray
\newcommand{\sYes}{\textcolor{ScoreYes}{\rule[-0.4pt]{6pt}{6pt}}}
\newcommand{\sPart}{\textcolor{ScorePart}{\rule[-0.4pt]{6pt}{6pt}}}
\newcommand{\sNo}{\textcolor{ScoreNo}{\rule[-0.4pt]{6pt}{6pt}}}

\title{Beyond Masks: The Case for Medical Image Parsing}

\author{
Siddharth Gupta\textsuperscript{1,2} \quad
Alan L. Yuille\textsuperscript{1} \quad
Zongwei Zhou\textsuperscript{1,3}\thanks{Correspondence to: Zongwei Zhou (\href{mailto:zzhou82@jh.edu}{\textsc{zzhou82@jh.edu}})} \\[2.5mm]
\textsuperscript{1}Johns Hopkins University \quad
\textsuperscript{2}Northwestern University \quad
\textsuperscript{3}Johns Hopkins Medicine
}

\begin{document}

\maketitle

\begin{abstract}
Medical imaging research has spent a decade getting very good at one thing: producing per-voxel masks. Masks tell us size, volume, and location, and a decade of clinical infrastructure rests on those outputs. Yet the report a radiologist writes contains almost nothing a mask can express. We argue that medical imaging research should adopt \emph{medical image parsing} as its central output: a structured representation in which \emph{entities}, \emph{attributes}, and \emph{relationships} are emitted together and mutually consistent. Entities are the named structures and findings, present or absent. Attributes describe those entities, capturing things like margin regularity, enhancement pattern, or severity grade. Relationships connect them, naming where one structure sits relative to another, what abuts what, and what has changed since the prior scan. A good parse satisfies three properties, in order: (1) \emph{decision} (the parse names the right things in the current image), (2) \emph{reconstruction} (its content is rich enough to regenerate that image), and (3) \emph{prediction} (its content is rich enough to forecast how the patient state will evolve). Quantitative measurements are derived from this content; they are not predicted alongside it. To test how close the field is to producing such an output, we audit eleven representative systems against the three parsing primitives plus closure. None emits a well-formed parse. Entities are largely solved. Attributes, relationships, and closure remain near-empty. The path forward is not a new architecture. It is a commitment to a richer output, and to training signals that reward it. Segmentation taught models to measure. Parsing asks them to explain.
\end{abstract}

\section{Introduction}
\label{sec:intro}

Medical imaging research has spent a decade getting very good at one thing: producing per-voxel masks. Masks tell us size, volume, and location, and a decade of clinical infrastructure rests on those outputs. Yet the report a radiologist writes contains almost nothing a mask can express. \emph{This paper asks one question: what output should medical imaging AI emit?}

A radiology report identifies structures, qualifies them with attributes such as margin regularity or steatosis grade, connects them with relationships such as \emph{located in segment VII} or \emph{abuts the pleura}, flags what is absent, and tracks what has changed since the prior scan. A segmentation mask encodes the extent of a structure and nothing else. The gap is not one of mask quality. Modern systems produce excellent masks. The gap is one of output representation: mask-only outputs cannot express attributes, relationships, absence flags, or temporal edges.

The case for parsing is positive, not only defensive. Clinicians are rate-limited reporters: clinical guidelines specify quantitative content radiologists are trained to assess but omit under time pressure. Tumor volume is more sensitive than the longest-axis number RECIST 1.1 substitutes for it \citep{eisenhauer2009recist} and trivial to compute from a mask, yet reports rarely contain it. A parser is not rate-limited. It can populate every guideline slot for every entity at every timepoint.

The framework that captures what is missing already exists. Tu, Chen, Yuille, and Zhu \citep{tu2005image} introduced \emph{image parsing} as the joint inference of segmentation, detection, and recognition over a structured output, arguing that the three tasks mutually improve each other when solved together. They formulated parsing as \emph{analysis by synthesis}, a classical idea in vision tracing back to Helmholtz, in which perception is the inverse of generation. To understand an observation, posit a structured cause that could have generated it, and verify by synthesis. We adopt this framework for medical imaging and extend it forward in time. A parser maps a medical observation to a structured patient state. A synthesis decoder closes the loop on the current observation by regenerating it from the state. A state-dynamics model extends the loop into the future by predicting how the state will evolve. A parse is good if it satisfies three properties: \emph{decision} (correct entities, attributes, and relationships), \emph{reconstruction} (regenerate the image through synthesis), and \emph{prediction} (forecast future state through dynamics). \S\ref{sec:parsing} makes all three precise.

Recent medical imaging work has begun moving toward parsing. BiomedParse \citep{zhao2025biomedparse} scales the entity primitive across 82 object types and nine modalities. Grounded report generators \citep{bannur2024maira2, tanida2023rgrg} attach language to regions. Image-to-graph systems \citep{xiong2023priorradgraphformer, jain2021radgraph} emit typed relations. None yet produces all three primitives together as a mutually consistent structured output, and none closes the analysis-by-synthesis loop. Figure~\ref{fig:parse-example} previews what such an output looks like.

\begin{figure}[t]
    \centering
    \includegraphics[width=\linewidth]{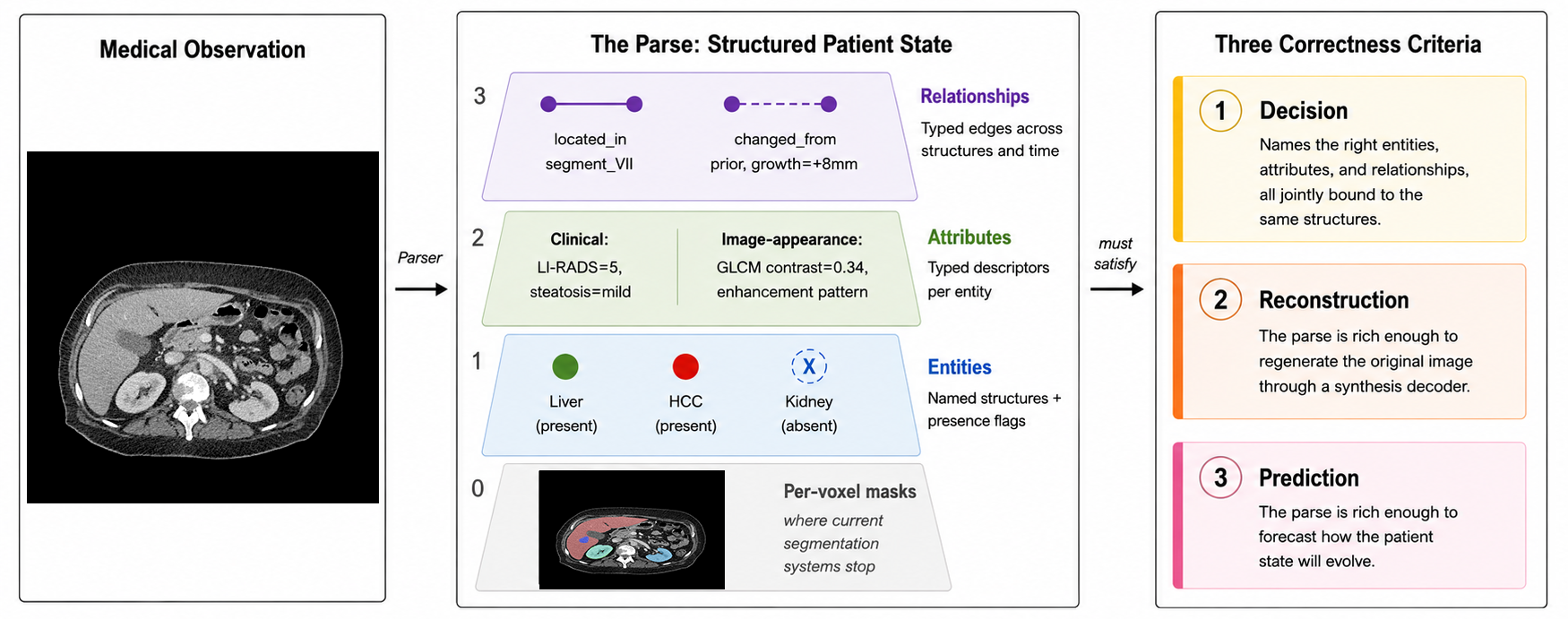}
    \caption{
    Medical image parsing extends segmentation by emitting a layered structured output. From a medical observation (left), a parser produces a parse composed of entities (named structures with masks and presence flags), attributes (clinical and image-appearance descriptors), and relationships (typed edges across spatial, semantic, and temporal classes). Per-voxel masks are the base layer where current segmentation systems stop; parsing adds three structured layers on top. The parse must satisfy three correctness properties (right): \emph{decision} (entities, attributes, and relationships are correct and jointly bound), \emph{reconstruction} (the parse is rich enough to regenerate the observation), and \emph{prediction} (the parse is rich enough to forecast future state). \S\ref{sec:parsing} formalizes each component.
    }
    \label{fig:parse-framework}
\end{figure}

We argue that medical image parsing will become the dominant frame for medical imaging research over the next decade. Medical image parsing extends segmentation by treating entities, attributes, and relationships as first-class outputs of a structured state that supports the right clinical decision on the current image, is rich enough to reconstruct it, and is rich enough to predict how the patient will evolve. The field should commit to this representation now, before the next generation of benchmarks locks the community into another decade of mask-only progress. Figure~\ref{fig:parse-framework} shows the full framework.

This paper makes three contributions:
\begin{itemize}
    \item A definition of \emph{medical image parsing} with three primitives (\emph{entities}, \emph{attributes}, \emph{relationships}) and a closure condition that binds them.
    \item Three correctness properties for a parse: \emph{decision} (the parse names the right things), \emph{reconstruction} (the parse content is sufficient to regenerate the current image through $p_{\text{syn}}$), and \emph{prediction} (the parse content is sufficient to forecast future state through $p_{\text{dyn}}$).
    \item A capability audit of eleven representative systems against the three parsing primitives plus closure, identifying entities as largely solved and attributes, relationships, and closure as near-empty.
\end{itemize}

\section{What segmentation cannot represent}
\label{sec:cannot}

Per-voxel labels are a lossy projection of any structured scene. We catalog four classes of content that vanish in the projection. Each pattern appears across computer vision; each is most consequential in medical imaging, where the next clinical action depends on it.

\begin{figure}[b]
    \centering
    \includegraphics[width=1\linewidth]{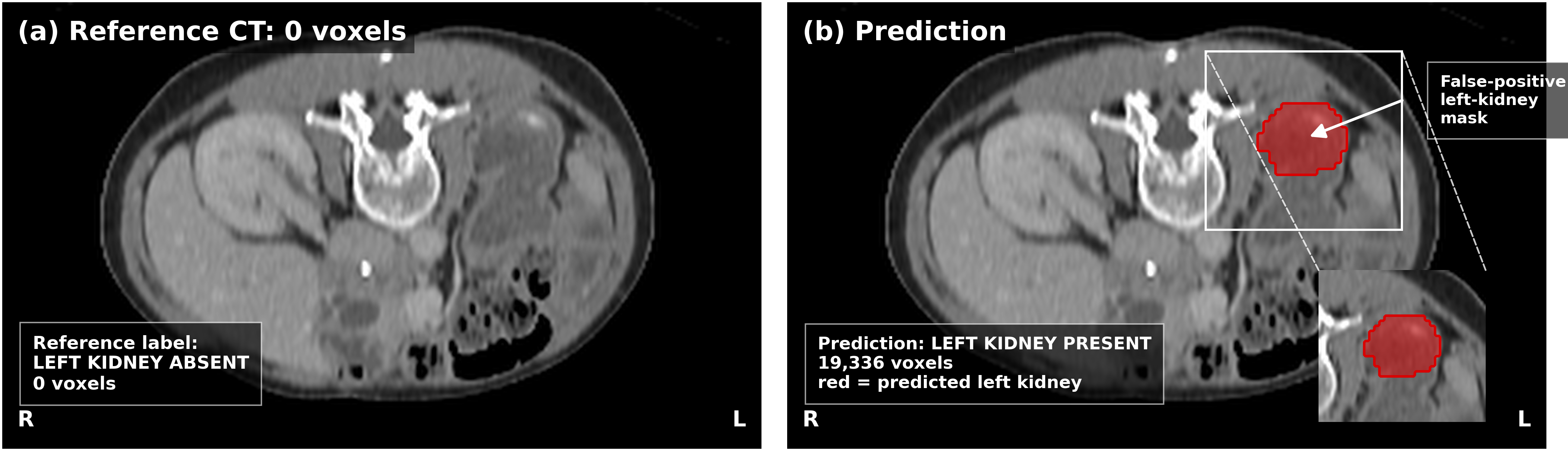}
    \caption{
    Example false-positive organ mask. In public CT case s1223, the reference left-kidney mask contains zero voxels, while a pretrained nnU-Net-based segmentor (TotalSegmentator-fast) predicts a left-kidney mask of 19{,}336 voxels. A mask-only model can produce an anatomically plausible organ mask even when the target organ is absent from the reference annotation.
    }
    \label{fig:absence-hallucination}
\end{figure}

\textbf{Absence is not a mask.} A segmentation model has no native way to declare an entity absent. Trained to label kidneys, it produces a spurious mask in the perinephric fat of a post-nephrectomy patient (Figure~\ref{fig:absence-hallucination}). Trained to label vehicles, it hallucinates cars in empty driveways. Absence is the non-existence of a region, outside the vocabulary of voxel-wise labeling. Detection heads emit ``no object''; segmentation outputs do not.

\textbf{Attributes are not classes.} Per-voxel labels cannot encode properties of structures. ``Red car,'' ``tilted face,'' and ``broken chair'' are properties of objects, not new object types; ``fatty liver,'' ``ground-glass opacity,'' and ``irregular margin'' are conditions of anatomical structures, not new structures. Encoding each as a separate class produces a combinatorial explosion no annotation budget can absorb. Attributes have no spatial extent independent of the structures they qualify.

\textbf{Relationships are not mask overlap.} Two equally accurate masks can share the same pixel layout while standing in very different relations. ``Person riding horse'' and ``person next to horse'' look identical at the mask level; the relation distinguishes them. Whether a hepatic mass sits in Couinaud segment VII or VIII changes the surgical approach \citep{strasberg2005nomenclature}; whether a pulmonary nodule abuts the visceral pleura changes its malignancy prior \citep{macmahon2017fleischner}. These facts cannot be read off whole-organ masks: AbdomenAtlas \citep{li2024abdomenatlas} labels the liver as one mask rather than eight Couinaud segments; PanTS \citep{li2025pants} labels the pancreas as one organ rather than head, body, and tail. Spatial relationships are bottlenecked by granularity; semantic and temporal relationships exceed masks regardless of granularity. \S\ref{sec:parsing} develops the full taxonomy.

\textbf{Change is not mask subtraction.} Subtracting two per-frame masks does not capture change, even when both masks are individually accurate. An oncology model that emits accurate liver-tumor masks at two MRIs does not say the tumor grew \citep{eisenhauer2009recist}. Pixel-wise mask subtraction conflates change with segmentation noise, registration error, and acquisition variability. Worse, masks cannot represent the kinds of change that matter clinically: a tumor whose mask is unchanged but whose interior has become heterogeneous is progressing under modern criteria; a liver whose surrounding parenchyma has fibrosed has changed in a way that affects resectability. Capturing such drifts requires representations that compare across time, not masks that subtract.

These four failures are consequences of the output representation, not of weak models or insufficient data. The remainder of the paper argues that the next decade should treat entities, attributes, and relationships as first-class outputs, with masks as one component rather than the whole.

\section{Medical image parsing}
\label{sec:parsing}

\S\ref{sec:intro} introduced medical image parsing as inference of a structured patient state from a medical observation. We now make this precise. Let $\mathbf{o}_t = \{\mathbf{I}_t, \mathbf{r}_t, \mathbf{c}_t\}$ be a medical observation at clinical encounter $t$ (images, reports, clinical variables), and let $\mathbf{s}_t = \{E_t, A_t, R_t\}$ be a structured patient state composed of \emph{entities}, \emph{attributes}, and \emph{relationships}. A parser maps observations to states under the distribution
\begin{equation}
\hat{\mathbf{s}}_t \sim p_{\text{perc}}(\mathbf{s}_t \mid \mathbf{o}_t),
\label{eq:parsing}
\end{equation}
which generalizes to $p_{\text{perc}}(\mathbf{s}_t \mid \mathbf{s}_{0:t-1}, \mathbf{o}_{0:t-1}, \mathbf{o}_t)$ when the patient's prior trajectory is available. Today's CT is read against prior scans, lab trends, and family history; a 3-year-old normal scan and a 3-month-old stable scan are different signals. A synthesis decoder closes the loop on the current observation:
\begin{equation}
\hat{\mathbf{o}}_t \sim p_{\text{syn}}(\mathbf{o}_t \mid \mathbf{s}_t),
\label{eq:synthesis}
\end{equation}
generating an observation from the parse so that perception can be checked against its own predictions. A state-dynamics distribution extends the loop forward in time:
\begin{equation}
\hat{\mathbf{s}}_{t+\Delta} \sim p_{\text{dyn}}(\mathbf{s}_{t+\Delta} \mid \mathbf{s}_t, \mathbf{o}_t, \mathbf{a}_t, \Delta),
\label{eq:dynamics}
\end{equation}
forecasting future state across a horizon $\Delta$, conditioned on any action $\mathbf{a}_t$ taken between $t$ and $t+\Delta$ (which may be ``no action,'' i.e., natural progression). It generalizes to full-history conditioning $p_{\text{dyn}}(\mathbf{s}_{t+\Delta} \mid \mathbf{s}_{0:t}, \mathbf{o}_{0:t}, \mathbf{a}_{0:t}, \Delta)$ when the patient's prior trajectory is available. Segmentation is the special case in which $p_{\text{perc}}$ has support only over per-voxel masks of $\mathbf{I}_t$. Parsing strictly extends segmentation in both input and output. The probabilistic framing is intentional: the same observation can support multiple plausible parses, and clinical software downstream needs calibrated confidence on each primitive. We adopt the entity/attribute/relationship vocabulary from scene-graph generation \citep{tu2005image} and clinical relation extraction \citep{jain2021radgraph, wu2021chestimagenome}. This paper focuses on \emph{image parsing}; reports and clinical variables enter as supporting context, not as the primary parsing target.

\textbf{\textit{Entities}} are the named things a clinician identifies: organs (liver, right kidney), substructures (Couinaud segment VII), and findings (a hepatocellular carcinoma, a ground-glass opacity). Each carries an identity that persists across timepoints, a segmentation mask when spatial delineation is meaningful, and a typed presence flag. An entity may be declared absent, resolving the failure mode that segmentation cannot express.

\textbf{\textit{Attributes}} qualify entities and split into two kinds. \emph{Clinical} attributes are categorical or ordinal labels radiologists assess against established guidelines: a margin is irregular, a hepatic lesion has LI-RADS category 5 \citep{chernyak2018lirads}, a liver has steatosis grade 2. They include the biomarkers that clinical guidelines specify for diagnosis, staging, and treatment selection. Their vocabularies should be drawn from the clinical decision frameworks the field already uses: LI-RADS, BI-RADS, Lung-RADS, PI-RADS, and oncology consensus guidelines (NCCN). \emph{Image-appearance} attributes are quantitative descriptors of image content within an entity: texture features, enhancement-curve parameters, attenuation values, and parenchymal-state descriptors of the surrounding organ. They capture content that the clinical guidelines do not name but downstream tasks rely on, such as the texture of a tumor's interior or the condition of the parenchyma surrounding a lesion. Hepatic resectability, for example, depends not only on tumor size but on the health of the remaining liver, since a cirrhotic remnant cannot regenerate the way a healthy remnant can. Attributes are typed and entity-scoped, which collapses the combinatorial explosion segmentation faces: ``fatty liver'' is not a new class but the attribute \texttt{steatosis=moderate} on the entity \texttt{liver}.

\textbf{\textit{Relationships}} connect entities within a single timepoint. \emph{Spatial} relationships (\texttt{located\_in}, \texttt{abuts}) describe geometric configuration. \emph{Semantic} relationships (\texttt{drains\_from}, \texttt{causes}) encode anatomical or causal facts not recoverable from spatial overlap, however fine: lymphatic drainage is not the lymph node closest to a primary tumor but the node in its drainage basin. Relationships are typed edges, not overlaps between masks.

\textbf{Temporal change.} The three primitives together support longitudinal reasoning without a separate temporal mechanism. When prior parses are available, they enter Eq.~\ref{eq:parsing} as conditioning input and make prior entity identities, attribute values, and relationship structure available to current perception; the same $E$, $A$, $R$ vocabulary applies at every timepoint. Change is then expressed structurally: a \texttt{changed\_from} edge between persistent entities, with a payload that records the difference (e.g., \texttt{growth=+8\,mm}, \texttt{steatosis: mild} $\to$ \texttt{moderate}, or \texttt{margin: smooth} $\to$ \texttt{irregular}). Because identity persists, the question ``did this tumor grow?'' is well-defined: \texttt{tumor\_001} in $\mathbf{s}_{t-1}$ and \texttt{tumor\_001} in $\mathbf{s}_t$ refer to the same finding, and the change is a structural difference between their fields rather than a difference between two masks.

A good parse must satisfy three properties: \emph{decision}, \emph{reconstruction}, and \emph{prediction}. Each forces additional content into the parse, and each is testable against a separate signal.

\textbf{The decision property.} The first property asks whether the parse names the right things. Each entity declared must correspond to a structure or finding actually present in the image; each attribute assigned must match what a radiologist would assess; each relationship drawn must be correctly typed. The parse must agree with grounded gold annotations on a joint-correctness metric: an entity-attribute-relationship triple is correct only if all three are correct \emph{and} bound to the same entity. Per-primitive correctness is necessary but not sufficient: a system that names an entity correctly, an attribute correctly, and a relationship correctly but binds them to the wrong combination of entities has not parsed the image. Decision is the property that grounded report generators \citep{bannur2024maira2, tanida2023rgrg} and scene-graph systems \citep{xiong2023priorradgraphformer, jain2021radgraph} target most directly.

\textbf{The reconstruction property.} The second property goes beyond naming. The parse $\mathbf{s}_t$ must be sufficient to regenerate $\mathbf{o}_t$ through Eq.~\ref{eq:synthesis}: any sample $\hat{\mathbf{o}}_t \sim p_{\text{syn}}(\mathbf{o}_t \mid \mathbf{s}_t)$ must be close to $\mathbf{o}_t$ under a task-conditioned clinical distance $d_{\text{clin}}(\mathbf{o}_t, \hat{\mathbf{o}}_t) \leq \epsilon$. The metric is sensitive to clinically meaningful features (tumor volume for prognosis, enhancement profile for LI-RADS, parenchymal attenuation for resectability) and tolerant of clinically irrelevant variation (photon noise, breathing phase, normal anatomical variability). The reconstruction need not be pixel-perfect; it must be good enough that no clinically relevant content is missing. This forces image-appearance attributes into the parse: masks and clinical labels alone are insufficient to regenerate the image. The property traces back to the original formulation of image parsing \citep{tu2005image} as Bayesian inversion of a generative model, and provides a deployment-time self-check: a hallucinated entity causes the synthesis to render content the input lacks, and the residual surfaces the parse error without ground-truth labels.

\textbf{The prediction property.} The third criterion goes beyond reconstruction. The parse must support not only regenerating \emph{this} observation but also forecasting \emph{the next}. Through the state-dynamics distribution Eq.~\ref{eq:dynamics}, a parse $\mathbf{s}_t$ rolls forward to a predicted future state $\hat{\mathbf{s}}_{t+\Delta}$, which the synthesis decoder renders into a predicted future observation $\hat{\mathbf{o}}_{t+\Delta} \sim p_{\text{syn}}(\mathbf{o}_{t+\Delta} \mid \hat{\mathbf{s}}_{t+\Delta})$. The prediction property requires $d_{\text{clin}}(\mathbf{o}_{t+\Delta}, \hat{\mathbf{o}}_{t+\Delta}) \leq \epsilon$ when the actual future observation $\mathbf{o}_{t+\Delta}$ becomes available. Reconstruction tests whether the parse captures clinically relevant content of the present; prediction tests whether the parse captures clinically relevant trajectory. The two are complementary, and a parse that wins on the first while failing on the second has missed something the future depended on. A parse that reconstructs but does not predict is a snapshot. A parse that predicts is a state.

\textbf{Closure.} A latent code $\mathbf{z}_t$ from an autoencoder also supports reconstruction, but it is opaque and not consumable by clinical software. A parse must be both reconstruction-supporting \emph{and} structurally typed: every attribute and every relationship endpoint must resolve to a declared entity. We call this requirement \emph{closure}; it is what distinguishes a parse from a latent code.

\begin{figure}[!t]
    \centering
    \includegraphics[width=\linewidth]{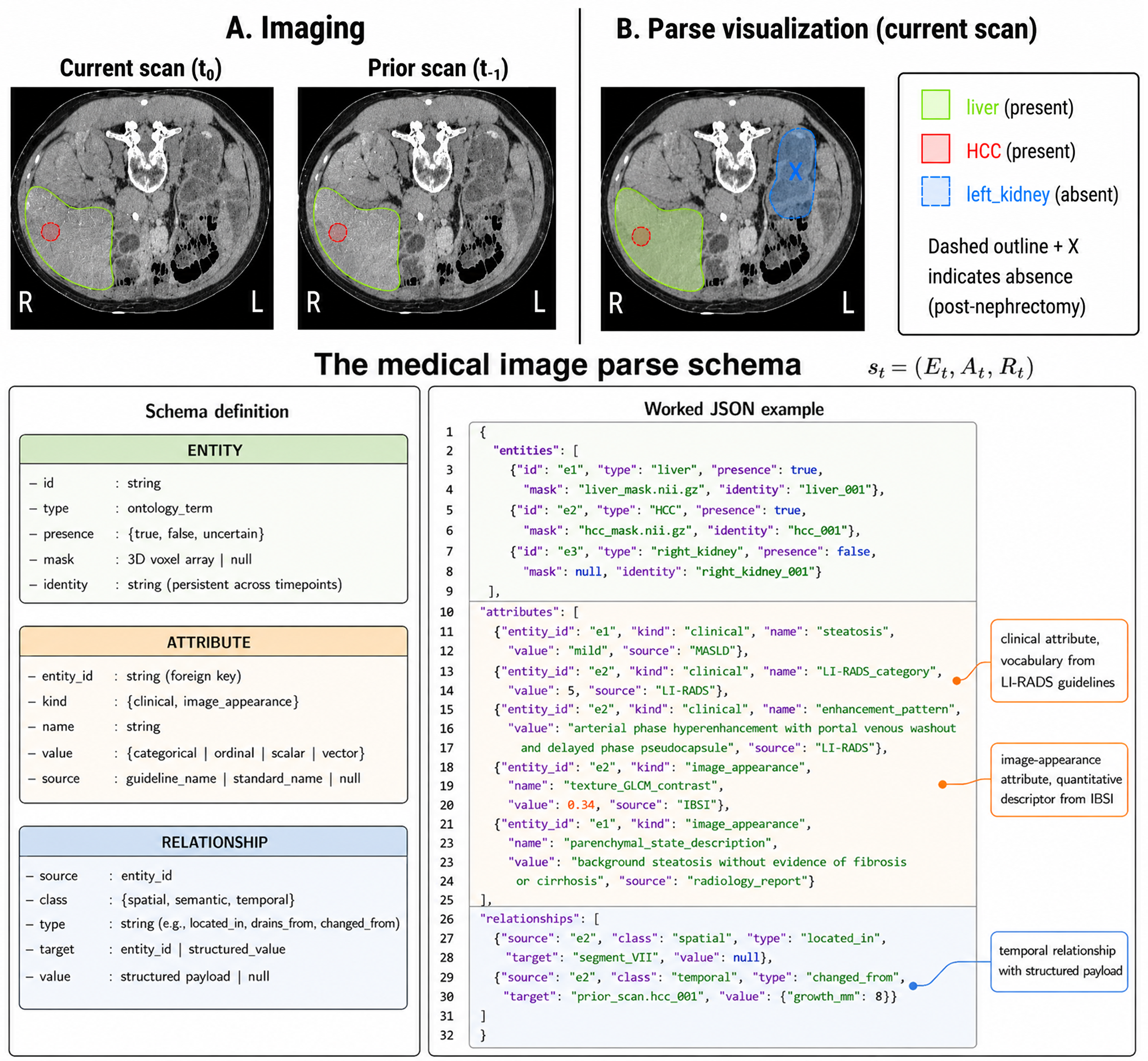}
    \caption{
    Example medical image parse. Two entities are present (\texttt{liver}, \texttt{HCC}) and one is absent (\texttt{left\_kidney}, post-nephrectomy). Attributes (clinical and image-appearance) qualify each entity; relationships (\texttt{located\_in}, \texttt{changed\_from}) connect them within and across timepoints. The structure is consumable by surgical planning software and rich enough to drive a synthesis decoder for reconstruction-based audit.
    }
    \label{fig:parse-example}
\end{figure}

Figure~\ref{fig:parse-example} illustrates a parse for a liver MRI with a prior scan. Quantitative measurements (diameter, volume, growth rate) are derived from the parse by deterministic computation, not predicted by the model: a correct mask already determines diameter, so asking a model to regress diameter alongside the mask asks it to relearn a function it already represents. Medical image parsing is not a new model. It is a commitment to an output representation, and to the three properties (decision, reconstruction, prediction) that make the representation more than a structured caption.

\section{A capability audit of current systems}
\label{sec:audit}

The definition in \S\ref{sec:parsing} admits a direct empirical question: how close are existing medical imaging systems to producing a well-formed parse? To answer it, we audit eleven representative systems across the camps that have approached structured output in medical imaging. \emph{Joint detection-segmentation-recognition} systems (BiomedParse \citep{zhao2025biomedparse}, Foundation X \citep{islam2025foundationx}, ME-VLIP \citep{shaaban2025mevlip}) produce multiple outputs per entity but treat each entity independently. \emph{Self-identified medical parsers} (Ghesu RL agents \citep{ghesu2016rlagent}, ME-VLIP) adopt the terminology of parsing but commit to a narrower output than \S\ref{sec:parsing} requires. \emph{Scene-graph approaches} (Prior-RadGraphFormer \citep{xiong2023priorradgraphformer}, building on the Chest ImaGenome schema \citep{wu2021chestimagenome}) produce typed relations but at the cost of spatial grounding. \emph{Grounded report generators} (MAIRA-2 \citep{bannur2024maira2}, RGRG \citep{tanida2023rgrg}) attach language to regions but leave the language itself as the attribute representation. \emph{Hierarchical and taxonomic segmentors} (SAT \citep{zhao2025sat}) scale entity vocabularies without introducing attributes or relations. Two further systems sit alongside these camps rather than within them: Adam-v2 \citep{taher2024adamv2}, a part-whole self-supervised representation showing that even structured-understanding efforts stop short of structured output; and CheXRelNet \citep{karwande2022chexrelnet}, the only system in the audit that emits longitudinal change as a typed edge. We score each system on the three primitives of \S\ref{sec:parsing} plus closure in Table~\ref{tab:audit}.

\begin{table}[t]
\centering
\scriptsize
\setlength{\tabcolsep}{3pt}
\renewcommand{\arraystretch}{1.25}
\caption{Capability audit of eleven medical imaging systems against \S\ref{sec:parsing}'s three primitives plus closure. Systems ordered by publication year. \sYes{} explicit, typed output; \sPart{} partial or implicit support; \sNo{} not supported. Absence handling is folded into Entities; temporal change is folded into Relationships. Reconstruction and prediction (the other two properties of \S\ref{sec:parsing}) are not shown as columns: no audited system supports either.}
\label{tab:audit}
\begin{tabular}{@{}lcccc@{\hskip 0.4em}lcccc@{}}
\toprule
System & Entities & Attributes & Relationships & Closure & System & Entities & Attributes & Relationships & Closure \\
\midrule
Ghesu RL Agent 2016        & \sPart & \sNo   & \sNo   & \sNo   & RadGraph 2021               & \sPart & \sYes  & \sYes  & \sPart \\
CheXRelNet 2022            & \sPart & \sPart & \sPart & \sNo   & RadGraphFormer 2023   & \sPart & \sYes  & \sYes  & \sPart \\
RGRG 2023                  & \sYes  & \sPart & \sNo   & \sNo   & Adam-v2 2024                & \sPart & \sNo   & \sNo   & \sNo   \\
MAIRA-2 2024               & \sPart & \sPart & \sNo   & \sNo   & BiomedParse 2025            & \sYes  & \sPart & \sNo   & \sNo   \\
Foundation X 2025          & \sYes  & \sPart & \sNo   & \sNo   & ME-VLIP 2025                & \sPart & \sPart & \sNo   & \sNo   \\
SAT 2025                   & \sYes  & \sNo   & \sNo   & \sNo   &                             &        &        &        &        \\
\bottomrule
\end{tabular}
\end{table}

Three patterns in Table~\ref{tab:audit} are worth reading carefully.

\textbf{The field has largely solved entities.} Four systems produce entities as explicit, delineated outputs with masks and presence information, and several others handle absence through invalid-prompt detection, negative-region flags, or certainty modifiers. This is the corner of parsing where a decade of segmentation, detection, and region-based reporting has accumulated genuine capability. BiomedParse in particular demonstrates that joint detection-segmentation-recognition can be scaled to 82 object types across nine modalities with explicit support for entities being absent. When reviewers of this paper ask what the field has achieved, this is the honest answer: the entity primitive is largely in place, and treating it as open is a mischaracterization.

\textbf{Attributes and relationships are where the field is empty.} Only two systems score \checkmark{} on attributes (Prior-RadGraphFormer and RadGraph), and both do so in text, not as typed properties of grounded entities. No image-level medical system in the audit emits typed, entity-scoped attributes as structured output. The picture is similar for relationships: only Prior-RadGraphFormer and RadGraph produce typed edges, and only by importing the schema from the Chest ImaGenome corpus. No system generates relationships from images end-to-end as a structured object consumable by clinical software. Temporal relations are the rarest sub-type. CheXRelNet \citep{karwande2022chexrelnet} alone treats progression as a typed edge between anchored regions; MAIRA-2 consumes prior images as input but does not emit change as an explicit relation. The shortfall on attributes and relationships is not a gap within one camp. It is a gap across all five.

\textbf{No system produces a well-formed parse, supports reconstruction, or supports prediction.} The Closure column is the starkest in the table: only Prior-RadGraphFormer and RadGraph earn partial scores (their text-level graphs satisfy internal closure within the report); every other system shows no closure support. The two columns we did not include in the table are bleaker still. No audited system takes a full parse as conditioning input to a synthesis decoder (the reconstruction property of \S\ref{sec:parsing}), and none uses a parse to drive a state-dynamics model (the prediction property). Across closure, reconstruction, and prediction, the field is essentially empty. This is not a failure of model capacity but of output commitment. Systems that approach parsing from the vision side (BiomedParse, Foundation X) emit entities but no binding structure; systems that approach it from the language side (MAIRA-2, RadGraph) emit binding structure but without image grounding sufficient for closure. The parse as an object does not currently exist in the literature.

Three systems sit closest to the definition of \S\ref{sec:parsing} and deserve explicit recognition. \textbf{BiomedParse} is the most advanced realization of the entity primitive at scale; its gap is that attributes, relationships, and closure are outside its objective. \textbf{Prior-RadGraphFormer} is the most advanced realization of attributes and relationships as structured output; its gap is that it operates on text graphs with only partial spatial grounding, producing neither masks nor a closure condition that binds to image evidence. \textbf{MAIRA-2} is the most advanced realization of grounded description; its gap is that attributes live in unstructured language and relationships are not emitted at all. These three systems represent the current frontier along three different axes of parsing, and the gap between them is not closable by integration alone: each was trained against a different objective. Closing the gap requires a shared output representation and a training signal that rewards closure. Both are research problems, not engineering ones.

\section{Four components of a parsing system}
\label{sec:components}

The audit in \S\ref{sec:audit} identified that no current system produces a well-formed parse, and \S\ref{sec:parsing} made explicit what such a parse must contain. Closing the gap requires four components, each operationalizing one piece of the framework in \S\ref{sec:parsing}.

\textbf{The parser, $p_{\text{perc}}$.} A perception model that emits all primitives jointly: entities with masks and presence flags, attributes typed and entity-scoped, relationships across spatial, semantic, and temporal classes, with prior-state conditioning when available. The closest current systems stop short along three different axes (joint detection-segmentation-recognition, scene graphs, grounded reports). The barrier is not architectural capacity but training objective: each was trained against one projection of parsing rather than the structured object containing all projections.

\textbf{The audit decoder, $p_{\text{syn}}$.} The synthesis distribution that operationalizes the reconstruction property. With it, parse correctness can be checked against the input image at deployment without ground-truth labels. Components exist piecewise (conditional generative models \citep{hu2023label, chen2024towards, lai2024pixel, li2026text, mao2025medsegfactory}, the Medical World Model line \citep{yang2025medical}, the synthesize-then-compare pattern of \citet{xia2020synthesize} and structure-consistency anomaly detection \citep{xiang2024exploiting}), but none takes a full parse as conditioning or trains against a clinical-distance objective.

\textbf{The dynamics model, $p_{\text{dyn}}$.} The forward extension of the parse: a generative model that rolls a parse trajectory (full history when available) to a future state under horizon $\Delta$ and action $\mathbf{a}_t$. It operationalizes the prediction property: a parse that wins on reconstruction but fails on prediction has missed something the future depended on. The closest precedent is the Medical World Model line \citep{yang2025medical}, which conditions diffusion decoders on partial state to simulate post-treatment evolution. None is conditioned on a full parse with multi-visit history, and none supports arbitrary horizons.

\textbf{The structural diff, $\Delta \mathbf{s}_t = \text{Compare}(\mathbf{s}_t, \mathbf{s}_{t-1})$.} The operation that surfaces change between two parses: identifies entities whose presence flag flipped, attributes whose values shifted, and relationships that appeared or disappeared. Because identity persists, the diff is well-defined. CheXRelNet \citep{karwande2022chexrelnet} treats progression as a typed edge between anchored regions across timepoints; alternatives such as MAIRA-2 leave comparison implicit in hidden state. Explicit diff is consumable by clinical software; implicit comparison is not.

\section{Alternative views}
\label{sec:alt}

The position is contestable along three axes: scaling, integration, and simplicity. We address each.

\textbf{Foundation models will learn parsing implicitly.} The opposing view holds that vision-language foundation models trained on image-report pairs at scale will produce attributes and relationships as a byproduct of next-token prediction. CLIP-style contrastive pretraining aligns image and text embeddings, not structured output, and is weak where parsing matters most (small-lesion localization); even purpose-built medical CLIP-driven systems \citep{liu2024universal} emit masks and detections, not typed attributes or relationships. More fundamentally, implicit knowledge is not auditable, reproducible, or consumable by clinical software, and recent benchmarks of medical VLMs on tumor-centric tasks \citep{chen2025vision} and audits of automated radiology report generation \citep{zhang2025automated} reach the same verdict. A model that says ``segment VII,'' ``segment 7,'' and ``anterosuperior segment of right hepatic lobe'' across three queries about the same finding has failed at producing a parse. A VLM redesigned around structured-output prediction is a parser by another name, which is this position, not a counter-position.

\textbf{Existing systems can be integrated rather than retrained.} A second view: stack BiomedParse for entities, Prior-RadGraphFormer for relationships, and MAIRA-2 for descriptions. But each was trained against a different objective, and stacked outputs do not enforce mutual consistency: BiomedParse's entity prediction and Prior-RadGraphFormer's relationship endpoint may not refer to the same object, and no signal in the pipeline detects the failure. The mutual-improvement principle (segmentation improves under relationship constraints, attributes improve under correct entity identity) requires joint training. Stacked predictions are post-hoc reconciliation, which is the problem closure exists to detect.

\textbf{Clinicians want decisions, not parses.} A third view: radiologists want triage outputs (``this lung nodule is suspicious, recommend biopsy''), not relational graphs. The objection conflates model output with user interface. A parse is what downstream software consumes, not what the clinician reads. Surgical planning software reads the parse and presents a 3D model with annotated segments. A binary triage decision is a projection of a parse, not a competitor. The reconstruction property strengthens this: the audit decoder renders an image from the parse, and the clinician compares input to parse-driven reconstruction rather than reading JSON. Radiologists already produce structured outputs daily; AI that emits less structure than the clinicians it supports is a step backward.

\section{Call to action}
\label{sec:call}

For the position to translate into measurable progress, the field needs shared infrastructure that lets parse-level work compose across groups. We name three calls to action: benchmarks, datasets, and algorithms. The schema for what a parse contains has been provided in \S\ref{sec:parsing} and Figure~\ref{fig:parse-example}; the calls below address what the field must build around it.

\subsection{Build benchmarks across decision, reconstruction, and prediction}

A parse supports three distinct uses, and one benchmark cannot test all three. We call for a benchmark triple, one per property of \S\ref{sec:parsing}. The \textbf{decision benchmark} scores the parse against grounded gold annotations using a joint-correctness metric: an entity-attribute-relationship triple is correct only if all three are correct \emph{and} bound to the same entity. RadFact \citep{bannur2024maira2} and RadGraph-F1 \citep{jain2021radgraph} began moving in this direction; a parse-level metric extends the principle across image-grounded primitives simultaneously. The \textbf{reconstruction benchmark} checks whether a synthesis decoder, conditioned on the parse, can regenerate the input under $d_{\text{clin}}$, extending the synthesize-then-compare pattern of \citet{xia2020synthesize} from segmentation to full parses. The \textbf{prediction benchmark} rolls the parse forward through Eq.~\ref{eq:dynamics} and scores the predicted future state against the actual one (tumor-volume trajectory, treatment response, new findings at follow-up). A parse rich enough to drive a forward model is a more rigorous standard than one rich enough to label a single scan; this benchmark connects parsing to the world-model line \citep{yang2025medical}. The three are independent: a parser can pass one and fail others, which is exactly the diagnostic information the field lacks. Touchstone \citep{bassi2024touchstone} is a precedent for the multi-axis benchmark this triple requires; recent audits of medical AI challenges \citep{lubonja2025auditing} reinforce that benchmark and metric design are themselves research problems.

\subsection{Annotate datasets covering all parse content}

Existing datasets cover one or two parse components at most: AbdomenAtlas \citep{li2024abdomenatlas, chen2025scaling} and PanTS \citep{li2025pants} contain masks; RadGraph \citep{jain2021radgraph} contains text-level entities and relations; Chest ImaGenome \citep{wu2021chestimagenome} contains region-level scene graphs; PadChest-GR \citep{castro2025padchestgr} contains grounded findings with progression labels. None contains, at meaningful scale, masks paired with typed entity-scoped attributes, typed relationships, presence flags, and temporal edges. The path forward is scalable annotation: pre-annotation by foundation models with clinical verification (as in ScaleMAI \citep{li2025scalemai}, RadGPT \citep{bassi2025radgpt}, and Label Critic \citep{bassi2025label}), report-derived supervision that substitutes for dense masks \citep{bassi2025scaling}, hierarchical annotation that reuses existing mask labels as anchors for attribute and relationship layers, and active learning that prioritizes informative cases. The dataset needs to cover all parse content end-to-end on enough cases to train and evaluate against the benchmarks above.

\subsection{Develop algorithms that produce parses jointly}

Benchmarks and datasets are inert without models trained against them. Today's systems were each trained against an objective that targeted one projection of parsing: segmentation masks, text-level relation graphs, or grounded sentences. Closing the gap requires algorithms whose training objective is the parse itself: entities, attributes, and relationships emitted jointly, with the reconstruction property of \S\ref{sec:parsing} enforced through paired synthesis training. The architectural choice (transformer, graph network, diffusion model) is left open. Whatever the architecture, the training signal must reward joint correctness across entities, attributes, and relationships rather than per-component accuracy on any single one.

These three calls are not separable. A benchmark without a dataset is empty leaderboard infrastructure. A dataset without an algorithm is supervision the field does not consume. An algorithm without a benchmark is a system no one can verify. Together, they constitute the minimum infrastructure for the position of this paper to translate into measurable progress.

\section{Conclusion}
\label{sec:conclusion}

The position of this paper is a bet about how research communities work: what gets measured determines what gets researched. The field has measured masks for a decade and has produced excellent mask producers. If it continues to measure only masks, it will continue to produce only mask producers, regardless of how much capacity is added or how much data is fed in.

The alternative is a paradigm the field has already begun reaching toward but not yet committed to. Joint detection-segmentation-recognition systems, scene-graph models, and grounded report generators are each pushing past mask-only outputs along a single axis. None yet emits a complete parse, and none will until the benchmarks, datasets, and algorithms exist that make parse-level performance the unit of progress.

What a community measures is what it learns to produce. The field has taught itself to produce masks. It can teach itself to produce parses.

\begin{ack}
This work was supported by the National Institutes of Health (NIH) under Award Number R01EB037669. We would like to thank the Johns Hopkins Research IT team in \href{https://researchit.jhu.edu/}{IT@JH} for their support and infrastructure resources where some of these analyses were conducted; especially \href{https://researchit.jhu.edu/research-hpc/}{DISCOVERY HPC}.
\end{ack}

% \clearpage
\bibliographystyle{plainnat}
\bibliography{zzhou,refs}

\end{document}